\documentclass[11pt]{article}

\usepackage[final]{acl}

\usepackage{times}
\usepackage{latexsym}

\usepackage[T1]{fontenc}

\usepackage[utf8]{inputenc}

\usepackage{microtype}

\usepackage{inconsolata}

\usepackage{graphicx}

\newcommand{\sk}[1]{\textcolor{blue}{}}

\usepackage[utf8]{inputenc} 
\usepackage[T1]{fontenc}    
\usepackage{hyperref}       
\usepackage{url}            
\usepackage{booktabs}       
\usepackage{amsfonts}       
\usepackage{nicefrac}       
\usepackage{microtype}      
\usepackage{xcolor}         
\usepackage{amsmath,amssymb,amsthm} 
\usepackage{graphicx}       
\usepackage{algorithm}      

\usepackage{float}          
\usepackage{subcaption}     
\usepackage{booktabs}       
\usepackage{caption}        
\theoremstyle{definition}

\usepackage[most]{tcolorbox}
\newtcolorbox{outputbox}{
    colback=white, 
    colframe=gray!30,
    boxrule=0.5pt,
    arc=5pt,
    enhanced,
    breakable,
    pad at break=10pt,
    top=5pt,
    bottom=5pt,
    left=10pt,
    right=10pt,
    width=\linewidth,
    fontupper=\small\itshape
}

\usepackage{algpseudocode}
\title{Leveraging Pretrained Language Models as Energy Functions for Glauber Dynamics Text Diffusion}

\author{%
  Tarun Kathuria \\
  UC Berkeley\\
  \texttt{tarunkathuria@berkeley.edu} \\
  \And
  Sachin Kumar \\
  Ohio State University \\
  \texttt{kumar.1145@osu.edu} \\
}

\begin{document}

\maketitle

\begin{abstract}
 We present a discrete diffusion-based language model using Glauber dynamics from statistical physics. Our main insight is that instead of trying to train a discrete state space diffusion model using Glauber dynamics with a uniform transition kernel as the forward process, one can set up an ``energy function'' based on pretrained causal/masked language models. When viewed as the stationary distribution, this energy function allows us to significantly improve the quality of the generated text. Incorporating UL2 as the pretrained model into our diffusion pipeline, we outperform prior diffusion based LMs and perform competitively with autoregressive models of comparable model sizes. Furthermore, our models are competitive with or outperform prior diffusion models and GPT-2 style auto-regressive models on  zero-shot common sense reasoning tasks as well as planning and search tasks like Sudoku and Zebra puzzles. 
 \end{abstract}

\section{Introduction}
The dominant paradigm for training language models has been autoregressive (AR), where given the preceding context, models are trained to generate the next token in the sequence.
However, AR models face a number of challenges, especially in solving tasks involving global planning, complex structural constraints, and self-correction \citep{lin-etal-2021-limitations,pmlr-v235-bachmann24a,huang2024large}. Diffusion LMs are a promising alternative in addressing some of these limitations and unlocking new capabilities \citep{ye2024diffusion,zhang2023planner,tae2025tess} building on their success in continuous domains like image and video generation \citep{nichol2021improved,tae2022editts,ho2022video,ramesh-etal-2024-evaluating}. 
Continuous space diffusion models first define a forward Markov process that gradually converges to an easy-to-sample-from distribution (e.g., Gaussian). 
They then need to learn time reversal of this process, or denoising, that can transform the noise back into a clean data sample. 

Directly applying diffusion models to discrete text is nontrivial. Prior work has explored various approaches to either approximate text in a continuous domain  \citep{li2022diffusion,han2022ssd,richemond2022categorical} or modified the underlying diffusion mechanism to directly operate on discrete data~\citep{lou2024discrete,gong2024scalingdiffusionlm,lou2024discretediffusionmodelingestimating}. Existing discrete diffusion approaches, however, often suffer from instability, slow training, weak theoretical foundations, or inefficient sampling due to reliance on heuristic transition rules or approximations \citep{varma2024glauber}. Among these models, masked diffusion LMs (MDLMs) have received significant recent attention showing fast inference while almost matching generative quality of AR models, as well as improved performance on downstream reasoning benchmark \cite{kim2025train}.  
However, \citet{mdlm-limitation} rigorously demonstrate that \textbf{MDLMs ultimately are highly unlikely to outperform AR models}. 
These limitations of MDLMs stem from a fundamentally flawed forward Markov process and noisy/stationary distribution; we expand on this in  \autoref{sec:relWork}. Furthermore, 
\cite{diffusion_serializable} also show that certain problems are inherently non-parallelizable and highlight other key difficulties of MDLMs whose inherent strength is their parallel sampling at the cost of often poorer quality.

Instead, we propose a first-principles discrete diffusion approach based on Glauber dynamics, a well-studied Markov process in theoretical CS and statistical physics \citep{levin-peres}. Our main insights are that diffusion models are pathwise relative entropy minimizing \citep{follmer85, lehec13} and, hence, their performance depends heavily on how far the ``noisy distribution'' is from the data distribution as well as the behavior of the underlying stochastic dynamics (measured using a  notion of curvature). 

Glauber dynamics is a natural way to address the latter, typically being the dynamics of choice for sampling from many challenging discrete state space distributions. It provides a well-defined and easy-to-implement Markov chain often with guaranteed convergence to a stationary distribution, offering a principled method for generative modeling. 
Furthermore, Glauber dynamics is typically expected to shine in the presence of an ``energy function'', i.e., sampling from $p(x)\propto e^{-f(x)}$ for some $f$ operating on our discrete domain of interest. We propose to use pretrained LMs (such as AR or masked LMs) as our energy function, treating them as our ``noisy distribution'' allowing us to leverage a significant amount of compute effort spent on training them and speed up the training process for our diffusion LM. 
This choice also addresses the bottleneck of sample inefficiency of training diffusion LMs and poorer performance compared to autoregressive LMs as we would typically expect the sampling distribution of an AR model to be closer to the data distribution as opposed to a uniform or unigram distribution as used in prior works \citep{lou2024discretediffusionmodelingestimating, varma2024glauber}.
While we sketch how to use GPT-style models as energy functions in the appendix, we opt to use UL2 style models~\citep{tay2022ul2} as our energy functions, which are trained to span multiple tasks, including causal generation as well as mask infilling, as this provides a unifying framework under which our Glauber dynamics-based diffusion pipeline can be easily understood and implemented.

Starting with UL2 weights as initialization and adding additional variables for incorporating the temporal aspect of diffusion models, we train our model on the score-entropy loss function \citep{lou2024discretediffusionmodelingestimating, benton22} adapted to the forward Markov process defined by Glauber dynamics with an energy function. We obtain significantly better (unconditional) generative perplexities than all prior discrete diffusion LMs and are competitive with the generative and zero-shot perplexities of AR models. We also report improved performance on MAUVE scores \citep{pillutla2021mauve} as well as on some common-sense reasoning tasks like Winogrande \citep{ai2:winogrande}, PIQA \citep{Bisk2020}, SIQA \citep{sap2019socialiqa} and HellaSwag \citep{zellers2019hellaswag}. Very recent work also shows that if masked diffusion language models are trained to robustly account for token orderings, then they can outperform Autoregressive models of 7 times the parameter count on logic puzzles like Sudoku and Zebra puzzles \citep{kim2025train}. We show that without any changes to our training procedure to robustly account for token orderings, our UL2 Glauber dynamics can outperform MDLMs and hence AR models on the same Sudoku and Zebra puzzles tasks. Finally, we demonstrate that Glauber dynamics yields better \emph{quality per unit inference compute} than sampling from AR models in a compute-normalized best-of-$N$ experiment: using a T5-Gemma 2B--2B backbone \cite{t5gemma25} fine-tuned on reasoning and language tasks, a single Glauber-UL2 generation ($N=1$, costing $2L$ model invocations) matches or exceeds an AR model generating two independent candidates at the same compute budget, and the advantage grows further at $N=2$. This experiment also serves as a direct example of reusing a pretrained popular open source LLM to convert into a Glauber UL2 diffusion model with boosted performance. 

\section{Preliminaries}
\subsection{Discrete Diffusion Models}
Our goal is to model a probability distribution over sequences of length $L$ and hence our state space is $\Omega = \Sigma^L$ where $\Sigma$ is a finite vocabulary. 
A discrete diffusion process describes (time-varying) Markov chains $Q_t$ with probability distributions $p_t \in \mathbb{R}^\Omega$ evolving according to a discrete heat equation:
\begin{align*}
    \frac{dp_t}{dt} = Q_t p_t \ \ \ p_0 \approx p_\mathcal{D}
\end{align*}    
where $Q_t \in \mathbb{R}^{\Omega \times \Omega}$ are the generators of the Markov process   with non-negative off-diagonal entries and columns summing to zero and $p_{\mathcal{D}}$ is our data distribution. Furthermore, the process is assumed to be ergodic (and ideally fast-mixing) with a limiting distribution $p_\mathrm{base}$ as $t\rightarrow\infty$. The process is simulated via small Euler steps or scaled Poisson clocks \citep{campbell22}. 
Diffusion admits many different interpretations including one based on time-reversal of Markov processes \citep{anderson82} where we again set up a (necessarily time-varying) Markov process $\overline{Q}_t$:
\begin{align*}
\frac{d p_{T-t}}{dt} &= \overline{Q}_t \, p_{T-t} \\
\overline{Q}_t(y,x) &= e^{\log p_t(y) - \log p_t(x)} Q_t(x,y) \\
\overline{Q}_t(x,x) &= - \sum_{y \neq x} \overline{Q}_t(y,x)
\end{align*}
where the multiplicative factors $e^{\log p_t(y)- \log p_t (x)}=p_t(y)/p_t(x)$  are probability ratios which are the analogues of the score function in continuous space diffusion and are essentially the parameters to be learned. 

To train our model for reversing the process, we will use the score entropy loss framework of \citep{lou2024discretediffusionmodelingestimating}, where the ``denoising'' version of score entropy loss applied in the context of discrete diffusion models, called \textbf{diffusion weighted denoising score entropy} loss $\mathcal{L}_\mathrm{DWDSE}$ is defined as:
{\scriptsize
\begin{equation*}\label{eqn:dwdsel}
\begin{split}
\mathbb{E}_{x_0 \sim p_{\mathcal{D}}}
\Bigg[
\int_{0}^{T}
\mathbb{E}_{x_t \sim p_{t \mid 0}(\cdot \mid x_0)}
\sum_{y \sim x_t}
Q_t(x_t,y)
\Big(
    s_\theta(x_t,t)_y \\
\quad
    - \frac{p_{t \mid 0}(y \mid x_0)}{p_{t \mid 0}(x_t \mid x_0)}
      \log s_\theta(x_t,t)_y
    + K\!\left(
        \frac{p_{t \mid 0}(y \mid x_0)}{p_{t \mid 0}(x_t \mid x_0)}
      \right)
\Big)
\, dt
\Bigg]
\end{split}
\end{equation*}
}
    where $K(a) = a(\log a - 1)$ and $s_\theta (x,t)_y$ is the score function that is meant to be close to $p_{t\mid 0}(y)/p_{t\mid 0}(x)$ for all $x\in \Omega$ and all $y\sim x$, i.e. all neighbors $y\in \Omega$ of $x$.

This setup is general enough that once we fix the Markov chain and the corresponding stationary distribution, we can directly plug it into the loss for training our diffusion model. Hence, we focus our efforts on understanding what makes for good or bad Markov chains in discrete state spaces, especially for language modeling. 
As we discuss in \autoref{sec:relWork}, prior work shows that it is unlikely that currently popular masked diffusion LMs with absorbing Markov chain can surpass autoregressive LMs. We design a better alternative.

Instead of viewing diffusion models as a time reversal of a forward Markov process, we instead take the stochastic optimal control viewpoint of this process which is well-known to be the pathwise relative entropy minimizing stochastic process with end-point marginals corresponding to the data distribution and the stationary distribution of the corresponding Markov process \citep{follmer85,lehec13}. Given this insight, one desires that the stationary distribution is easy to sample from and it is efficient to make updates to a data point according to the (time-varying) Markov chains and the forward Markov chain. Furthermore, we desire that the data distribution is close in some sense to the stationary distribution (this ensures that one can converge in fewer steps to the data distribution starting from the stationary distribution). Another desirable property of the stochastic process, at least in the continuous diffusion case, is that some notion of curvature of the underlying stochastic process is sufficiently good \citep{conforti25} which typically manifests as the stochastic process having fast decay of entropy along the process trajectories. 

To account for the ask of the stationary distribution to be close to the data distribution, we first observe that a natural distribution which certainly should be closer to the data distribution than uniform/unigram distributions are that of pretrained language models. There are actually significant advantages of trying to use pretrained language models for diffusion models if possible. 
Unlike diffusion models, which receive one or a few gradient signals per step, AR models, trained using teacher forcing loss, obtain $L$ signals per step, facilitating more efficient learning of linguistic structure, which might be one reason why AR models perform so much better at learning the structure of language compared to diffusion models.

To address the issue of natural Markov chains, which likely have good underlying entropic decay, a natural suggestion is that of Glauber dynamics, which has been used widely in sampling from challenging distributions coming from theoretical computer science, statistical physics, and Bayesian inference. We essentially notice that the way Glauber dynamics samples at each step corresponds in some sense to masked language models and furthermore, we can use pretrained language models as energy functions or equivalently, conditional samples which will directly fit in quite naturally into the design of the forward process for Glauber dynamics. 

It is important to remark at this point that most prior discrete diffusion approaches like \citet{lou2024discretediffusionmodelingestimating} use Markov chains which let them sample many tokens in each iteration and hence their sampling is very fast, typically much faster than auto-regressive models which necessarily take $L$ model invocations to generate text. In our discrete diffusion model based on Glauber dynamics, generation will typically be slower than AR models (in our experiments it's $2L$ and $4L$ model invocations) which we acknowledge as an acceptable limitation especially because our focus is on generating much better quality text, which these prior masked diffusion language models struggle to do \citep{mdlm-limitation} and furthermore, for tasks which require search or planning, Glauber dynamics inherently seems better suited than either AR models or masked diffusion language models as the concept of natural back-tracking to edit incorrect tokens is naturally baked into the process. 

We remark that while our inference time is necessarily slower than auto-regressive and masked diffusion language models, there are ways to improve our inference time that might also lead to a reduction in our training time. In particular, there is work on parallelizing Glauber dynamics \citep{parallel-glauber} that can be leveraged here leading to improved sampling complexity, similar to work of \citep{paradigms} for continuous diffusion models for image generation.
\subsection{UL2 Models}
UL2 models form the backbone of our proposed Glauber diffusion models. The UL2 (Unified Language Learning) framework \citep{tay2022ul2} consolidates multiple pre-training objectives into a single versatile model.
Based on an encoder-decoder transformer architecture, UL2 is pretrained using a 

varied set of denoising tasks, where the model learns to reconstruct original text from corrupted versions. These tasks include R-denoising (regular span corruption, for general knowledge acquisition), S-denoising (sequential PrefixLM or sequence-to-sequence, for causal generation capabilities), and X-denoising (extreme span corruption, for recovering large missing portions of text).

While one could use AR models as energy functions in Glauber dynamics and we sketch one such way to do so, the flexibility of UL2 models to do both causal generation and mask infilling in a unified framework is instrumental to an efficient implementation of the diffusion transformer backbone in our Glauber dynamics based diffusion model. 

\section{Glauber Dynamics Discrete Diffusion}
In this section, we describe the training and inference procedure for our discrete diffusion model based on Glauber dynamics. We first describe the Glauber dynamics Markov chain to sample from a distribution on sequences of length $L$ with each index taking values in a discrete set $\Sigma$, where we are interested in sampling from a probability distribution $p:\Sigma^L \rightarrow [0,1]$. 

The sampling is broken up into $n$ different rounds where in each round, we a priori pick a permutation of $\{1,\ldots,L\}$ denoted by $\pi_i$ for $i=1,\ldots, n$ and sequentially update all $L$ tokens by updating just the $k=\pi_i(j)^{th}$ index in the $j^{th}$ iteration in the $i^{th}$ round according to the distribution $p(x_k = \cdot \ \vert x_{\setminus k})$, i.e., we sample $x_k$ from the stationary distribution conditioned on all but $x_k$ (denoted as $x_{\setminus k}$ indices being fixed. If $p(x) = e^{-f(x)}$ for some energy function $f$, the conditional distribution, with a Metropolis filter, can be  written as $p(x_k = \sigma \ \vert x_{\setminus k}) = \min\left\{1,e^{-f(x_{\setminus k}, \sigma) + f(x)}\right\}$, possibly with some self-loops to ensure lazy chains.  Even if the stationary distribution is not specified upfront and we just have the conditional distributions, one can show that under some mild conditions, there is a stationary distribution and the Glauber dynamics, which just needs these conditional distributions to run the sampler converges to this distribution. We remark that Glauber dynamics is often described as picking a random index to update, in each iteration, conditioned on the remaining indices being unchanged. However, in our setup, we will fix the permutations for each round upfront and for each data point, in the $t^{th}$ timestep, the same index is updated. As the time reversed Markov chain at each time-step has the same edge structure (up to reversal of edges) as the forward Markov chain at that time-step, in the reverse process during sampling, we will update the same index.

While our focus will be on a setup using encoder-decoder UL2 models that we describe next which is more efficient and forms a more coherent picture, we note that one can in principle use GPT-style causal models as a way to devise an energy function for the forward Glauber dynamics as well which we describe in appendix D. 

\subsection{UL2 based Glauber Dynamics}
We now describe our UL2 based forward process for Glauber dynamics. As described above, UL2 is a language model which is trained on both causal generation tasks as well as mask infilling tasks while sharing the same set of parameters. The first insight is that at each step of Glauber dynamics, we are essentially asking the model to take a specific index and conditioned on all other indices, asking it to give a probability distribution for what to fill this token with. So it is rather easy to see that this is essentially a mask infilling task! Furthermore, to understand the stationary distribution, it would essentially correspond to running our masked language model iteratively over each token multiple times. However this can be rather slow but given that in UL2, the masked language model shares the same weights for the causal generation task, it's a reasonable idea to just consider a causal generation from the UL2 model as an (approximate) sample from the stationary distribution which is what we will do during inference from the model.

Before describing training, we first describe our model architecture, especially in light of wanting to reuse pretrained model weights. However, the default UL2 model doesn't really have any time-varying aspect baked into it which is required for diffusion models. We convert a pretrained UL2 model $\theta$ into a diffusion transformer \citep{peebles-xie23} by adding time-embedding parameters $\gamma$, initialized to zero at $t=T$, but these time parameters will of course change as we train the diffusion transformer. 

\begin{algorithm}[H]
    \caption{UL2 Based Glauber Dynamics Diffusion Transformer Training }
    \label{alg:trainingAlgo}
    \begin{algorithmic}
        \Procedure{Training}{$\mathcal{D}$ dataset, $N$ number of diffusion rounds}
            \State DiT UL2 model $(\theta, \gamma)$ initialized with pretrained weights for $\theta$
            \For {Epoch $k \gets 1$ to $K$}
                \State Deep Copy $(\theta_k,\gamma_k)(\cdot, t=T) \rightarrow T_k$
               \For {Each iteration}
                    \ \State Sample $x \in \mathcal{D}$ and $t \in [0,T]$
                    \ \State From $x$, get $x_t $ by running $T_k$ as the forward process for $t$ iters (Details in Appendix A)
                    \State Compute SEDD Loss using $(\theta_k,\gamma_k)$ and $T_k$ on $x$ and $x_t$  (Details in Appendix A)
                    \State Update $(\theta_k,\gamma_k)$
                \EndFor
            \EndFor
            \State \Return{$(\theta_K,\gamma_K)$}
        \EndProcedure
    \end{algorithmic}
\end{algorithm}

During training, we will consider the UL2 model conditioned at time $T=N \times L$ and create a copy of these parameters because they are meant to serve as the Markov transition kernel as well as corresponding to the noisy distribution and not to be backpropagated on in our diffusion score entropy loss. This Markov transition kernel is then used to noise our data points up to time $t$ picked randomly and then fed into the Diffusion Score Entropy loss. The important thing to note is that our UL2 model already outputs a probability distribution that predicts filling a masked token with whereas in the score entropy loss function we are trying to learn the multiplicative changes in the transition probabilities $s_\theta(x\rightarrow(y=(x_{\setminus i}, y_i),t)$ corresponding to $p_t(y)/p_t(x)$ but notice that given the Markov transition kernel of the forward process, which is just the frozen copy of the UL2 model with time fixed to be $T$, we can just get the ratios by dividing by the probabilities of the UL2 model at time $t$. After one finishes a single epoch of training (we actually do this multiple times within an epoch to expedite training), one then discards the frozen copy of the UL2 model at time $T$ and considers the new updated model for creating the frozen copy. 

Our entire training algorithm is described in Algorithm \ref{alg:trainingAlgo} with some specific details about the forward process as well as how the SEDD loss can be computed just by the mask infilling probability distribution at time $T$ as well as at $t$ deferred to Appendix A. It is, however, important to note that because the model weights are shared for the causal generation and mask infilling task as well as across time, the frozen copy of weights that we're using to noise our data points is also changing. Hence the distribution for causal generation as well as mask infilling at the time endpoint of $T=N \times L$ is going to change over the course of the training.

\begin{algorithm}[H]
    \caption{UL2 Glauber Dynamics Inference }
    \label{alg:inferenceAlgo}
    \begin{algorithmic}
        \Procedure{Inference}{}
            \State DiT UL2 model $(\theta, \gamma)$, $N$ number of diffusion rounds, permutations $\pi_1,\ldots,\pi_N$
            \State $x\leftarrow $ [CAUSAL-GEN] using $(\theta,\gamma)$ invoked at fixed time $t=T$
            \For {$ n \gets N$ to $1$}
                \For {$ i \gets L$ to $1$}
                    \State $j \gets \pi_n(i)$\\
                   \State  Update $x_j$ via [MASK-INFILL] using $(\theta,\gamma)$ at time $t=n\times i$ on $(x_{1:j-1}$ [MASK] $x_{j+1:L})$
                \EndFor
            \EndFor
            \State \Return{$x$}
        \EndProcedure
    \end{algorithmic}
\end{algorithm}

Next, we describe the unconditional inference algorithm. Specifically, we take our trained UL2 based diffusion transformer model and given the $N$ permutations that were used during training, we first generate our $L$ tokens by invoking the CAUSAL-GEN mode from our model at time $T=N\times L$. Following that, for $N$ rounds in reverse, we update the tokens in reverse order of the corresponding permutation by repeatedly calling our UL2 diffusion transformer in MASK-INFILL mode, at different values of time corresponding to which token is to be updated at that time step.

We remark that the above algorithm describes unconditional generation. If a prefix is conditioned to be some string $PFX$, then we just freeze those indices and do causal generation on the tokens following the prefix and then during the Glauber dynamics Mask-Infilling steps, skip the token indices corresponding to the prefix. 

Also, like \citet{lou2024discretediffusionmodelingestimating,varma2024glauber}, our Diffusion Transformer architecture uses best practices like AdaLN-Zero and RoPE for the time and  positional embeddings respectively.

\section{Experiments}

\subsection{Setup}

\paragraph{Datasets and Models}
To enable a fair comparison with \citet{varma2024glauber} without retraining their setup, we train our models on OpenWebText \citep{owt19} and consider model architectures of similar size as GPT-2 Medium (350M) and GPT-2 Large (745M). Our UL2 models are trained following the procedure described in \citep{tay2022ul2} which uses Flan-T5 architecture and instruction following procedure \citep{flan-t524} and then trained on the mixture-of-denoisers objective, following which we add approximately 15\% more parameters for the time variables. We would ideally have wanted to use a pre-trained UL2 model, however, the released model checkpoint was only of 20 billion parameters and we could not find a checkpoint for a smaller model size. Instead, for our larger model, we simply take the FLAN-T5-LARGE model from Hugging Face and train it on the mixture of denoisers objective to get the corresponding UL2 model we denote as UL2-large. Our custom UL2-medium uses a T5 architecture (16 encoder/decoder layers, $d_{model}=1024$, $d_{ff}=2048$) trained on the mixture-of-denoisers objective. We will use these model checkpoints as a baseline by doing causal generation before we attach any time variables and train on a score entropy loss referred to as UL2 pre-SEDD CAUSAL-GEN. 

After we attach the time variables and train these UL2 models according to the score entropy loss in Glauber dynamics, since the parameters are shared for mask infilling and causal generation, the probability distribution for causal generation corresponding to the end-point time of $N \times L$ (the noisy distribution) would have changed compared to pre-SEDD (Here, $N$ is the number of rounds in Glauber dynamics, i.e., the number of times each token is touched after doing a causal generation). Hence, we also add causal generation from this model as an additional baseline denoted as UL2 post-SEDD CAUSAL-GEN at time $N \times L$ for $N=3$. Finally, we have our Glauber dynamics based UL2 models trained on the score entropy loss and we report results for the two model sizes for $N=1$ and $N=3$ denoted as Glauber-UL2 post-SEDD and the total number of time-steps/end-point time, which is $N\times L$. Other baselines include GPT-2-M/L/XL, as well as a continuous diffusion model for text \citep{gulrajani2023likelihood} and MDLM \citep{sahoo24} and SEDD Absorb \citep{lou2024discretediffusionmodelingestimating} as well as GGM \citep{varma2024glauber}.

\textbf{Hyperparameters} We consider $L=1024$, the number of tokens in the generation and $N$, the number of Glauber dynamics rounds to be $1$ and $3$ bringing our inference steps (number of model invocations) to be $2048$ and $4096$ (coming from the 1024 causal generation steps followed by $1024 \times N$ mask infilling steps). Again to maintain a fair comparison to GGM, we evaluate the perplexity of our $1024$ samples using large GPT models (GPT2 Large, XL, Neo). We describe  our training hyperparameters, optimizer, compute setup and other details in Appendix D.

\subsection{Perplexity Results}
We summarize our main results for generative perplexities in \autoref{tab:genppl}. 
We achieve significantly better generative perplexity results than all prior diffusion based text generative models and furthermore match GPT-2-M and are very close to matching that of GPT-2-L. We also compare how causal generation from UL2 prior to fine-tuning on the SEDD loss as well as the UL2 model at time $T$ after fine-tuning the UL2 model on the SEDD loss and see that both those models lag in perplexity compared to the corresponding GPT-2 models illustrating the importance of running the reverse Glauber dynamics procedure to obtain our improved results. It is, however, interesting to see that the model perplexities for causal generation from UL2 models post SEDD improve relative to pre SEDD suggesting that training on score entropy via Glauber dynamics does improve the language understanding and generative capabilities of the UL2 model even if used as a causal generative model. Furthermore, while doing one round of reverse Glauber improves the performance slightly, it improves significantly after 3 rounds of Glauber dynamics and we believe it is possible for performance to improve and surpass GPT-2 with more rounds of Glauber dynamics. 

We also report in \autoref{tab:zsppl} the zero-shot perplexities of our models and some baselines using some of the datasets from \citep{radford2019language}.
\subsection{Iso-Compute Best-of-$N$ Comparison}
A natural concern with Glauber dynamics-based generation is that it requires more model invocations than a single autoregressive (AR) pass. However, in many modern inference pipelines---such as RL/GRPO-style sampling, test-time training, and best-of-$N$ selection---a fixed inference budget is already spent generating multiple candidates. In these settings, the appropriate comparison is not single-shot latency but rather \emph{quality per unit total inference compute}. For our Glauber-UL2 model with $N=1$, each generation consists of one AR pass ($L$ steps) followed by one full edit pass over all tokens ($L$ steps), totaling $2L$ model invocations---the same compute as an AR model generating two independent candidates.

To study this directly, we train a Glauber-UL2 model using a pretrained T5-Gemma 2B--2B encoder-decoder backbone \citep{t5gemma25}, augmented with time-embedding parameters and fine-tuned on the Glauber dynamics score-entropy objective with $N=1$ (i.e., $T=L$). Both the Glauber-UL2 model and the AR baseline (the same T5-Gemma backbone without Glauber augmentation) are fine-tuned, for one full epoch, on a mixture of GSM8K, Winogrande \citep{ai2:winogrande}, PIQA \citep{Bisk2020}, SIQA \citep{sap2019socialiqa} and OpenWebText \citep{owt19}, and evaluated on the respective test sets. For each prompt, we generate $2K$ candidates from the AR baseline and $K$ candidates from Glauber-UL2, matching total inference compute at roughly $2KL$ model invocations. The best candidate is selected under each task's evaluation criterion, with $K \in \{1, 2\}$.

\autoref{tab:bon} reports results under this iso-compute protocol. At $K=1$, Glauber-UL2 already matches or exceeds the AR baseline at $K=2$ on all three tasks, and at $K=2$ it outperforms AR at $K=4$ across the board. These results support the view that iterative self-correction via Glauber dynamics yields better return on inference compute compared to simply sampling more AR candidates.

\begin{table}[h]
\centering
\setlength{\tabcolsep}{3pt}
\resizebox{\columnwidth}{!}{%
\begin{tabular}{|l|c|c|c|c|}
\hline
\textbf{Task} & \textbf{AR (BoN=2)} & \textbf{AR (BoN=4)} & \textbf{Glauber (BoN=1)} & \textbf{Glauber (BoN=2)} \\
\hline
GSM8K & 43.9 & 46.1 & 46.9 & \textbf{50.4} \\ \hline
Winogrande & 68.3 & 69.7 & 69.4 & \textbf{71.2} \\ \hline
PIQA & 77.6 & 79.9 & 79.1 & \textbf{80.8} \\ \hline 
SIQA & 48.9 & \textbf{50.3} & 49.5 & 50.2 \\ \hline
\end{tabular}
}
\caption{Iso-compute best-of-$N$ accuracy (\%). AR BoN=$2K$ and Glauber BoN=$K$ match total inference compute at $\sim\!2KL$ model invocations. Both models use a T5-Gemma 2B--2B backbone fine-tuned on the same data mixture.}
\label{tab:bon}
\end{table}

\subsection{Additional Experimental Results}
\begin{table}[h!]
\centering
\footnotesize 
\setlength{\tabcolsep}{2.5pt}
\begin{tabular}{|l|c|c|c|c|}
\hline
\textbf{Model} & \textbf{LAMB.} & \textbf{WT2} & \textbf{WT103} & \textbf{1BW} \\
\hline
GPT-2-M & \textbf{15.60} & 22.76 & 26.37 & 55.72 \\
\hline
SEDD-M * & $ 42.77$ & $31.04$ & $ 29.98$ & $ 61.19$ \\
\hline
Glauber-UL2-M (N=1) * & $ 17.89$ & $ 23.95$ & $ 30.21$ & $ 56.12$ \\
\hline
Glauber-UL2-M (N=3) * & $ 17.14$ & $\mathbf{ 20.98}$ & $\mathbf{ 25.47}$ & $\mathbf{52.18}$ \\
\hline\hline
GPT-2-L & 10.87 & \textbf{19.93} & 22.05 & 44.58 \\
\hline
Glauber-UL2-L (N=1)* & $ 11.25$ & $ 21.54$ & $ 24.71$ & $ 47.62$ \\
\hline
Glauber-UL2-L (N=3) *& $\mathbf{10.14}$ & $20.35$ & $\mathbf{20.83}$ & $ \textbf{44.12}$ \\ \hline
\end{tabular}
\caption{Zero-Shot Validation Perplexities (LAMB: LAMBADA, WT: WikiText). * marked model perplexities are upper bounds}
\label{tab:zsppl}
\end{table}

In Table \ref{tbl:mauve}, we report the  MAUVE score \citep{pillutla2021mauve} of our model and compare it against the baselines from \citep{lou2024discretediffusionmodelingestimating}. Here, the generation happens conditioned on a fixed prefix. Both our Glauber dynamics based models (N=1,3) achieve better MAUVE scores than that of SEDD-medium and GPT-2-medium. 

\begin{table}[h]
\centering
\small
\begin{tabular}{l|l|l}
\hline
Method & Annealing & Mauve \\ \hline
GPT-2-M & Nucleus-0.95  & 0.955 \\
      & None & 0.802 \\ \hline
SSD-LM & Logit Thresh-0.95 & 0.919 \\ 
       & None & 0.312 \\ \hline
SEDD-M & None & 0.957 \\ \hline
Glauber-UL2-M (N=1) & None & 0.959 \\
Glauber-UL2-M (N=3) & None & 0.966 \\ \hline
\end{tabular}
\caption{MAUVE scores (higher is better)}
\label{tbl:mauve}
\end{table}

Next,  Table \ref{tab:csr} contains the performance of our model on some common sense reasoning tasks like Winogrande \citep{ai2:winogrande}, PIQA \citep{Bisk2020}, SIQA \citep{sap2019socialiqa} and HellaSwag \citep{zellers2019hellaswag}. These tasks involve prompting the model with multiple-choice questions (2–4 options) and evaluating accuracy based on the answer. Appendix B and C contain Sudoku/Zebra puzzle results and text generation examples, respectively.

\begin{table}[h]
\centering

\setlength{\tabcolsep}{2pt} 
\resizebox{\columnwidth}{!}{
\begin{tabular}{|l|c|c|c|c|}
\hline
\textbf{Model} & \textbf{HS} & \textbf{Wino} & \textbf{PIQA} & \textbf{SIQA} \\
\hline
GPT-2-M & 38.3 & 50.7 & 67.4 & 37.7 \\ \hline
SEDD-M & 31.5 & 49.0 & 56.1 & 35.4 \\ \hline
\citep{diffarllm}-M & 37.2 & 52.6 & 59.6 & 39.0 \\ \hline 
Glauber-M(N=1) & 37.4 & 49.9 & 66.8 & 37.4 \\ \hline
Glauber-M(N=3) & 40.5 & 52.9 & 68.9 & 39.1 \\ \hline
\end{tabular}
}
\caption{Performance of models on Common Sense Reasoning Tasks (HS: HellaSwag, Wino: Winogrande)}
\label{tab:csr}
\end{table}

\begin{table*}[ht]
\centering
\resizebox{\textwidth}{!}{ %
\begin{tabular}{@{}llcccc@{}}
\toprule
\multicolumn{3}{c}{\textbf{Evaluation Model}} & \textbf{GPT2-L (774M)} & \textbf{GPT2-XL (1.6B)} & \textbf{GPT-NEO (2.7B)} \\ \midrule
\textbf{Evaluated Model} & \textbf{Sampling Algorithm} & \textbf{Total Params} & \textbf{Gen. PPL ($\downarrow$)} & \textbf{Gen. PPL ($\downarrow$)} & \textbf{Gen. PPL ($\downarrow$)} \\ \midrule
\multicolumn{6}{c}{\textbf{Autoregressive Models}}\\ \midrule
GPT2-M \citep{radford2019language} & top-$p$, $p=0.8, L = 1024, T = 1024$ & $345$M & $12.4$ & $13.0$ & $14.5$ \\ \midrule
GPT2-L \citep{radford2019language} & top-$p$, $p=0.8, L = 1024, T = 1024$ & $774$M & $-$ & $6.5$ & $7.4$ \\ \midrule
GPT2-XL \citep{radford2019language} & top-$p$, $p=0.8, L = 1024, T = 1024$ & $1.6$B & $-$ & $-$ & $6.8$ \\ \midrule
\multicolumn{6}{c}{\textbf{Prior Diffusion Models}}\\ \midrule
Plaid \citep{gulrajani2023likelihood} & $\tau=0.9$, $L=1024, T=4096$ & $1.3$B & $19.7$ & $19.7$ & $17.9$ \\ \midrule
SEDD-M\citep{lou2024discretediffusionmodelingestimating} & $L=1024, T=2048$ & $424$M & $27.3$ & $28.0$ & $25.2$ \\ \midrule
MDLM \citep{sahoo24} &  $L=1024, T=1000$ & $170$M & $44.2$ & $45.4$ & $40.9$ \\ \midrule
GGM \citep{varma2024glauber} & top-$p$, $p=0.8, L = 1024, T = 4096$ & $387$M & $19.5$ & $19.9$ & $18.0$ \\ \midrule
\multicolumn{6}{c}{\textbf{UL2 and our UL2 DiTs}}\\ \midrule
UL2-M (pre-SEDD CAUSAL-GEN) & $L=1024$ & $368$M & $21.7$ & $21.3$ & $20.4$ \\ \midrule
UL2-M (post-SEDD, T=$3\times L$ CAUSAL-GEN) & $L=1024$ & $419$M & $19.1$ & $19.6$ & $19.9$ \\ \midrule
Glauber-UL2-M (post-SEDD, T=$L$, i.e., $N=1$) & $L=1024$ & $419$M & $17.1$ & $17.5$ & $16.6$ \\ \midrule
Glauber-UL2-M (post-SEDD, T=$3\times L$, i.e., $N=3$) & $L=1024$ & $419$M & $13.2$ & $13.7$ & $14.9$ \\ \midrule \midrule
UL2-L (pre-SEDD CAUSAL-GEN) & $L=1024$ & $783$M & $-$ & $14.2$ & $14.9$ \\ \midrule
UL2-L (post-SEDD, T=$3\times L$ CAUSAL-GEN) & $L=1024$ & $898$M & $-$ & $11.4$ & $11.5$ \\ \midrule
Glauber-UL2-L (post-SEDD, T=$L$, i.e., $N=1$) & $L=1024$ & $898$M & $-$ & $9.5$ & $9.9$ \\ \midrule
Glauber-UL2-L (post-SEDD, T=$3\times L$, i.e., $N=3$) & $L=1024$ & $898$M & $-$ & $6.9$ & $7.8$ \\ \midrule
\bottomrule
\end{tabular}
}
\caption{Generative Perplexities evaluated over $1024$ unconditional generations. We do not evaluate a larger model (e.g., GPT2-XL) with a smaller model (e.g., GPT2-M).}
\label{tab:genppl}
\end{table*}

\section{Related Work}\label{sec:relWork}

\paragraph{Continuous Diffusion for Text Generation} Early continuous diffusion approaches for text focused on Gaussian diffusion in latent embedding spaces ~\citep{li2022diffusion,gulrajani2023likelihood} or the vocabulary simplex \citep{han2022ssd,han2023ssd2,tae2025tess}. However, these generally underperform autoregressive models without extensive empirical tuning. The ones that do match or outperform them rely on heavy annealing and empirical alterations~\citep{han2022ssd}.

\paragraph{Discrete Space Markov Chain Based Text Diffusion} This line of work considers the space to be fixed length sequences in the discrete space 
corresponding to the text vocabulary (or 256 different pixel values in the context of discrete diffusion for images) and designs forward Markov chains directly in this space \citep{sohl2015deep}.
In the context of text diffusion, most papers consider discrete time, discrete space Markov chains with transitions corresponding to each token index being treated independently and for each index, each token is uniformly transitioning to any other token \citep{ho2020denoising,nichol2021improved,austin2021structured}.
The reverse Markov process is then a series of time-varying Markov chains. 
However, their text diffusion results still underperform significantly compared to autoregressive models. 

Some recent work has also attempted to devise a framework for score matching in the discrete space akin to continuous diffusion \citep{song-score-2021} with the denoising score entropy framework \citep{lou2024discretediffusionmodelingestimating,benton22} emerging as a promising approach especially in the context of text diffusion models with a 150M parameter model matching the generative perplexities of a similar parameter size GPT-2 model. Also, MDLM \citep{sahoo24} simplifies the D3PM setup with a simpler training objective designed for the specific masked diffusion model Markov chain and obtains seemingly better results. However, most prior discrete diffusion models use independent forward transition kernels. This independence means token indices updated during reverse sampling may not align with those from training, potentially undermining the benefits of the diffusion process’s time-varying nature. Indeed \citep{mdlm-limitation} formally show that for the masked diffusion model, the optimum model of the loss function in this context is actually equivalent to a (time-invariant) masked language model thus making it hard to justify masked diffusion models as a compelling alternative to autoregressive language models. Furthermore, they show that at lower floating-point accuracy, these masked diffusion language models can do a kind of temperature hacking which makes their perplexities look better than they really are and these values degrade significantly once these models are evaluated at 64-bit floating-point precision. Similarly, \cite{diffarllm} derive diffusion models from AR weights via attention-mask annealing. However, as they remain within the MDLM paradigm and use a comparable loss function, they likely inherit the fundamental limitations of MDLMs identified by \cite{mdlm-limitation}. Furthermore, their generative perplexity, when measured using GPT-2-large, results are worse than PLAID \cite{gulrajani2023likelihood} which our model outperforms and so we do not compare it for its generative quality. An exception to this line of MDLM (and modifications) however, which is also relevant to our work, is concurrent work of \citep{varma2024glauber} where they also propose a discrete diffusion model based on Glauber dynamics. They consider the noisy distribution as the unigram distribution of the data corpus and their training and inference pipeline is modeled by considering a token at each timestep and considering whether that token is a prediction due to noise or an actual signal. They frame training as $O(L)$ binary classification problems, outperforming SEDD at the 350M scale. Crucially, this binary framework restricts output complexity to $O(T|\Sigma|)$, avoiding the $O(T|\Sigma|^2)$ requirement of modeling full token-to-token transitions at every step. Despite a different setup, our UL2-based Glauber dynamics similarly scales as $O(T|\Sigma|)$ rather than $O(T|\Sigma|^2)$.

\section{Conclusion}
We presented a text diffusion model based on Glauber dynamics. By framing Glauber dynamics as a mask-infilling task and utilizing pretrained language models as energy functions, we developed a diffusion framework that, for the first time, matches the performance of GPT-2-medium and GPT-2-large in language modeling, and achieves state-of-the-art performance on complex search tasks like Sudoku/Zebra puzzles.

\section{Limitations}
Our approach is slower than prior discrete diffusion models and autoregressive (AR) baselines, as it requires more model invocations; however, this trade-off is justified by superior text quality, matching or outperforming GPT-2 models, unlike most prior MDLMs. Our work also requires significant computation resources for training. As we outline in an appendix, we needed 32 H100 GPUs running for a little under 6 days to train our larger model which is a significant time (however the GGM paper \citep{varma2024glauber} reports taking 8 days on TPU compute which in an iso-TFLOPs comparison corresponds to 24 H100s on a GPT-2-M model size as opposed to our training time on GPT-2-L model size). We do believe that this can be improved using many engineering optimizations that we did not pursue as well as the possibility of using Kronecker factorization based second order optimizers like Muon \citep{muon-llm} which achieve significant improvements over AdamW for training AR LLMs. Furthermore, it might be possible to use flow matching versions of our energy based Glauber dynamics model which may be easier to train and perform better as their training seems to be more stable in the case of continuous space diffusion as well as for masked text language models \citep{flowmatch23,dfm24,genmatch24} and we leave this for future work.

\nocite{sudoku-zebra}
\bibliography{custom,anthology}

\newpage
\appendix

\section{Training Algorithm Additional Details}
In this section, we provide more details about our training algorithm in words. We reiterate that our noisy distribution is the causal generation from the DiT UL2 model with T = 1024 or 3072, the idea is that this corresponds to causal generation from the initial UL2 model but of course that distribution has changed somewhat since the weights are shared in the DiT UL2 model. For the training procedure, we will only need the mask infilling capabilities of this DiT UL2 model (across all times). 

Now, at each epoch $k$, we freeze a copy of the model weights corresponding to that of the final timestep, T and denote that as $T_k$, this will essentially be our forward Markov chain used for noising samples. In actuality, within each epoch, every few iterations, we instead update the frozen copy of the model weights to be an exponential moving average of the frozen copy of the model weights and the current model weights in order to stabilize training. Now, in each iteration, we take a sample from our dataset $x$ and sample uniformly a time between $0$ and $T$. We then run $t$ steps of the forward Markov chain coming from $T_k$. As the Markov chain is sequential, in order to reuse many intermediate steps to get more gradient signals, we actually consider the $x_t$ not just for $t$ but for 32 timesteps between 0 and $t$ uniformly, i.e., $x_{i\times \lfloor{t/32}\rfloor}$. Now for each of these $x_t$'s we will compute the SEDD loss. 

Recall that the SEDD loss is the expectation over $x_0 \sim p_\mathcal{D}$ of
\begin{equation}
\begin{aligned}
\int_{0}^{T}
\underset{x_t \sim p_{t\mid 0}(\cdot \mid x_0)}{\mathbb{E}}
&\Bigg[
\sum_{y \sim x_t} Q_t(x_t,y)\Big(s_\theta(x_t,t)_y \\
&- \frac{p_{t\mid 0}(y\mid x_0)}{p_{t\mid 0}(x_t\mid x_0)} \log s_\theta(x_t,t)_y \\
&+ K\!\left(\frac{p_{t\mid 0}(y\mid x_0)}{p_{t\mid 0}(x_t\mid x_0)}\right)
\Big)
\Bigg]\, dt
\end{aligned}
\end{equation}
Now, for each term inside the integral, we understand how we can compute it to be used during backprop. The first term is $Q_t (x_t,y) s_\theta (x_t,t)_y$ which is essentially the probability of our model at $t$ in the reverse process so this is essentially the DiT model outputs at time $t$ so we can compute that directly. Now for the second term, to compute $\log s_\theta(x,t)_y$ we just write it as the first term $Q_t (x_t,y) s_\theta (x_t,t)_y$ (so computed as before) divided by the prediction according to $T_k$ our frozen forward noising model. Also, for computing $\frac{p_{t\vert 0}(y\vert x_0)}{p_{t\vert 0}(x_t \vert x_0)}$, we use the frozen model $T_k$ by storing the probabilities as we ran the Markov chain from $0$ to time $t$ (by computing the probabilities up to time $t-1$ and then taking the $t^{th}$ step for getting to $y$ versus getting to $x_t$). Since this is using the frozen model, there is no backpropagation on this.

\section{Sudoku and Zebra Puzzles Results}
In this subsection, we report our results for solving Sudoku and Zebra/Einstein puzzles\cite{sudoku-zebra}. In the Sudoku puzzle setup, we are given a $9\times 9$ grid where each cell is to be occupied by a number in the range $\{1, 2, \ldots , 9\}$. The constraints are that the numbers along each row and column should be unique. In addition, the numbers within each $3 \times 3$ mini-grid should also be unique. Given a set of initially filled positions, the goal is to figure out the values that can occur in the unfilled cells. In
standard Sudoku puzzles, there will always only exist a unique solution to the puzzle.

Our models are trained by taking a random but fixed permutation upfront for each round (including the initial auto-regressive generation) whereas the auto-regressive baseline is trained both with and without correct ordering information and the masked diffusion language model baseline results are taken directly from \cite{kim2025train} and we refer to that paper for more details. We stress that picking a random ordering upfront rather than the correct ordering information handicaps our model relative to the AR baseline and we believe this to be reasonable given that we're allowed to do multiple passes to correct our model outputs. We do not just mask-unmask one cell at a time during the refinement process and instead at any given time mask the 6 next cells in the permutation and predict just one cell at a time. This is because we observe much slower convergence when we were only masking-unmasking one cell at a time likely because it gets stuck in locally stuck configurations for the puzzles. We present our results in Table 5.

\begin{table*}[t] 
    \centering
    
    \begin{minipage}{0.48\textwidth}
        \centering
        \begin{tabular}{|c|c|c|}
        \hline
        \textbf{Model} & \textbf{Params} & \textbf{Accuracy} \\
        \hline
        AR (w/o ordering) & 42M & 9.73\%   \\
        \hline
        AR (w/ ordering) & 42M & 87.18\%  \\
        \hline
        MDLM (vanilla) & 6M & 6.88\%  \\
        \hline
        MDLM (Top-K Prob.) & 6M & 18.51\% \\ \hline
        MDLM (Top-K Margin) & 6M & 89.49\% \\ \hline
        Glauber-UL2 (N=1) & 7M & 42.26\%\\ \hline
        Glauber-UL2 (N=3) & 7M & 91.82\% \\
        \hline
        \end{tabular}
        \caption{Performance on Sudoku tasks}
        \label{tab:sudoku}
    \end{minipage}
    \hfill 
    \begin{minipage}{0.48\textwidth}
        \centering
        \begin{tabular}{|c|c|c|}
        \hline
        \textbf{Model} & \textbf{Params} & \textbf{Accuracy} \\
        \hline
        AR (w/o ordering) & 42M & 80.3\%   \\
        \hline
        AR (w/ ordering) & 42M & 91.2\%  \\
        \hline
        MDLM (vanilla) & 19M & 76.9\%  \\
        \hline
        MDLM (Top-K Prob.) & 19M & 98.5\% \\ \hline
        MDLM (Top-K Margin) & 19M & 98.3\% \\ \hline
        Glauber-UL2 (N=1) & 21M & 94.1\%\\ \hline
        Glauber-UL2 (N=3) & 21M & 98.7\% \\
        \hline
        \end{tabular}
        \caption{Performance on Zebra riddles}
        \label{tab:zebra}
    \end{minipage}
\end{table*}

Next, we present the Zebra or so-called Einstein riddles. These are a more verbal style of a puzzle where we need to fill in values in a grid again but this time the type of possible constraints is much richer. An example of such a riddle is as follows: 

There are 3 people next to each other in a row. Everyone has a different name: Ali, Rose, Randy. Every one
lives in a different colored house: gold, silver, indigo. Everyone likes a different drink: orange juice, beer,
coffee. Match the people to the correct value for each of their characteristics using the clues.
\begin{enumerate}
\item The person who likes orange juice is immediately to the left of the person who likes coffee.
\item The person who likes beer is somewhere to the left of the person who lives in the indigo house.
\item  The person at the 1st position is Rose.
\item  Randy is not the person who likes orange juice.
\item  Randy is the person who lives in the gold house.
\end{enumerate}
We present the accuracy results in Table 6. Ordering for our models and baseline comparison setup is the same as in the Sudoku case, however, here we don't need to keep 6 cells masked but rather just one at a time works fine. 
\section{Samples generated from our model}

In this section we give some unconditional and some prefix-conditioned examples from our larger model. 
\subsection{Unconditional Samples}
\textbf{Example Generation 1}
\begin{outputbox}
    Urban gardening has emerged as a powerful movement with deep historical roots and a promising future, reshaping how people interact with their environments and food systems. The practice of cultivating plants in urban spaces is not a new phenomenon; it has existed in various forms for centuries, serving as a vital source of nutrition, community, and resilience. In ancient civilizations, city-dwellers often grew crops within or just outside city walls to ensure a steady food supply during times of siege or scarcity. The famous Hanging Gardens of Babylon, whether myth or reality, symbolize the enduring human desire to bring nature into urban settings. Throughout history, urban gardens have provided fresh produce, medicinal herbs, and a sense of connection to the natural world. During the Industrial Revolution, as cities expanded and populations grew, many people lost access to land and traditional farming, leading to increased food insecurity and health problems. In response, urban gardening initiatives began to take shape, particularly in Europe and North America. Community gardens, allotments, and school gardens became common, offering city residents a way to grow their own food, improve their diets, and build social bonds. In the early 20th century, urban gardening gained momentum as a response to economic hardship. The Victory Gardens of World War I and II encouraged citizens to grow food to support the war effort and reduce pressure on national food supplies. These gardens not only contributed to food security but also fostered a sense of patriotism and community spirit. In the decades that followed, urban gardening experienced periods of decline and resurgence, influenced by social, economic, and environmental factors. By the late 20th century, concerns about industrial agriculture, food safety, and environmental degradation led to a renewed interest in urban gardening. The rise of the environmental movement and increased awareness of climate change further fueled the popularity of urban agriculture. Today, urban gardening takes many forms, from rooftop gardens and balcony planters to community gardens and urban farms. These spaces provide fresh, locally grown food, reduce the environmental impact of food transportation, and help mitigate the urban heat island effect. Urban gardens also serve as important educational tools, teaching children and adults about the origins of food, the importance of biodiversity, and the benefits of sustainable living. In many cities, urban gardening has become a tool for social change, empowering marginalized communities to take control of their food supply and create green spaces in underserved neighborhoods. Community gardens often become hubs for cultural exchange, bringing together people from diverse backgrounds to share knowledge, traditions, and harvests. The future of urban gardening looks bright, as advances in technology and design make it easier and more efficient to grow food in cities. Innovations such as vertical farming, hydroponics, and aquaponics allow for year-round cultivation, even in small or unconventional spaces. Smart irrigation systems, automated lighting, and soil sensors help urban gardeners optimize plant growth and conserve resources. Urban gardening is also being integrated into urban planning and policy, with cities around the world implementing green infrastructure initiatives and supporting urban agriculture projects. These efforts not only enhance food security but also contribute to climate resilience, biodiversity conservation, and public health. As cities continue to grow and face challenges such as climate change, population growth, and resource scarcity, urban gardening will play an increasingly important role in creating sustainable, livable, and resilient urban environments. The movement is supported by a growing network of organizations, activists, and policymakers who recognize the multiple benefits of urban agriculture. Educational programs, workshops, and online resources make it easier than ever for people to get involved and start their own gardens. The rise of social media has also helped to spread ideas, share success stories, and connect urban gardeners across the globe. Looking ahead, urban gardening has the potential to transform cities into greener, healthier, and more equitable places. By reimagining unused spaces, fostering community engagement, and embracing innovation, urban gardening can help address some of the most pressing challenges of our time. Whether on a balcony, rooftop, or vacant lot, every garden contributes to a larger vision of sustainable urban living. The history and future of urban gardening remind us of the power of small actions to create meaningful change, and the importance of nurturing our connection to the earth, even in the heart of the city. As more people embrace urban gardening, the movement will continue to grow, inspiring new generations to cultivate not only plants but also hope, resilience, and a sense of belonging in an ever-changing world.
\end{outputbox}

\newpage

\textbf{Example Generation 2}

\begin{outputbox}
    There’s something truly special about rainy days, something that just makes you want to stay inside and watch the drops fall against the window. Rainy days are cozy, they’re comforting, and they make you appreciate the little things in life, like a warm cup of tea or coffee, maybe a good book or a favorite movie. The sound of rain on the roof is soothing, almost hypnotic, and it can make you feel like you’re wrapped in a blanket of calm, even if you’re just sitting at your desk or lying on the couch. Rainy days slow things down, they make the world outside seem a little quieter, a little softer, and that’s a feeling you don’t get on sunny days. When it rains, people tend to stay indoors, and that can make the city feel different, more relaxed, less rushed. You might notice that everyone moves a little slower, talks a little softer, and seems to take their time. There’s a sense of unity on rainy days, as if everyone is sharing the same experience, the same rhythm of the weather. Rainy days are also great for creativity, or at least that’s what a lot of people say. Maybe it’s because there are fewer distractions, or maybe it’s because the rain itself inspires a certain mood, but rainy days seem to be perfect for writing, painting, or just thinking deeply about things. Sometimes, when it rains, you might find yourself daydreaming more, letting your mind wander to places you haven’t thought about in a while. Rainy days can be nostalgic, too, bringing back memories of childhood, like splashing in puddles or watching cartoons on a lazy afternoon. For some people, rainy days are a time for reflection, a chance to pause and take stock of life, to think about where you are and where you want to go. Rainy days can also be a little sad, but in a good way, like the kind of sadness that feels cleansing, that helps you let go of things you’ve been holding onto. The rain washes away the dust and the dirt, and sometimes it feels like it washes away your worries, too. Of course, not everyone loves rainy days. Some people find them depressing, or boring, or just inconvenient, especially if they have to go outside and get wet. But for those who do enjoy them, rainy days are a gift, a chance to slow down and appreciate the world in a different way. Rainy days are also great for naps, because the sound of the rain is like nature’s lullaby, and it’s easy to drift off to sleep with that gentle patter in the background. Sometimes, when it rains, you might notice the smell of the earth, that fresh, earthy scent that comes after a good downpour. It’s called petrichor, and it’s one of those little details that make rainy days special. Rainy days are also a good excuse to stay in your pajamas all day, or to wear your favorite sweater, or to make soup or bake cookies. There’s something about the combination of rain and comfort food that just feels right, like the universe is giving you permission to take it easy. Rainy days can also be romantic, especially if you’re sharing them with someone you love. There’s nothing quite like cuddling up with someone while the rain falls outside, or watching a movie together under a blanket, or just talking for hours without any distractions. Rainy days are a reminder that it’s okay to slow down, to take a break, to let yourself relax and recharge. In a world that’s always rushing, always demanding more, rainy days are a chance to pause, to breathe, to just be. Sometimes, when it rains, you might feel a little more introspective, a little more thoughtful, and that’s not a bad thing. Rainy days can help you reconnect with yourself, with your thoughts and feelings, and with the people around you. They can also be a time for new beginnings, because the rain brings new life, new growth, and new possibilities. After the rain, the world looks different, brighter, fresher, as if it’s been renewed. That’s one of the things I love most about rainy days—the way they make everything feel new again. Rainy days are also a great time for writing letters, or sending messages to friends, or just reaching out to someone you haven’t talked to in a while. There’s something about the rain that makes you want to connect, to share your thoughts and feelings, to let someone know you’re thinking of them. Rainy days can be a little lonely, too, but 

\end{outputbox}

\newpage

\textbf{Example Generation 3}

\begin{outputbox}
    sometimes I just forget things. Like, I’ll walk into a room and forget why I went in there. It happens all the time, almost every day, and it’s kind of funny but also a little annoying. I think everyone forgets things sometimes, like where they put their keys or what they were supposed to buy at the store. It’s just part of being human, I guess. Forgetting things can be frustrating, especially when you really need to remember something important, but it’s also kind of normal. I remember one time I forgot my friend’s birthday, and I felt really bad, but they said it was okay, and we laughed about it. Forgetting things is just something that happens, and it’s not always a big deal. Sometimes, when I forget something, I try to retrace my steps in my mind, like thinking about where I was or what I was doing before I forgot. That sometimes helps, but not always. Sometimes, I just have to accept that I forgot and move on. Forgetting things can be a little embarrassing, especially if it happens in front of other people, but most people understand because it happens to them too. I think forgetting things is just part of life, and it’s not something to worry about too much. Of course, if you forget things all the time, like really important things, then maybe it’s a good idea to talk to someone about it. But for the most part, forgetting things is just a normal part of being alive. Sometimes, I wonder why we forget things. Maybe it’s because our brains are too full, or maybe it’s because we’re distracted, or maybe it’s just because we’re tired. I think being tired makes you forget things more easily, at least that’s what I’ve noticed. When I don’t get enough sleep, I forget things more often, like where I put my phone or what I was supposed to do that day. Forgetting things can be frustrating, but it’s also kind of funny when you think about it. Like, how can you forget something you were just thinking about a minute ago? It’s weird, but it happens. Sometimes, I forget things because I’m thinking about something else, or because I’m in a hurry, or because I’m just not paying attention. I think paying attention is really important if you want to remember things, but it’s hard to pay attention all the time. Life is busy, and there’s always something going on, so it’s easy to forget things. Forgetting things is just part of being human, and it’s not something to be ashamed of. I think most people forget things sometimes, even if they don’t admit it. Sometimes, I forget things and then remember them later, like in the middle of the night or when I’m doing something completely unrelated. It’s funny how that works. The other day, I forgot where I put my glasses, and I looked everywhere for them, and then I found them on my head. That was pretty funny, and I laughed about it. Forgetting things can be annoying, but it can also be kind of funny, especially when you realize how silly it is. Sometimes, I forget things because I’m stressed or worried about something else, and my mind is just somewhere else. I think stress makes you forget things more easily, at least that’s what I’ve noticed. When I’m stressed, I forget things all the time, like appointments or deadlines or what I was supposed to do. It’s frustrating, but it’s also kind of normal. I think forgetting things is just part of being alive, and it’s not something to worry about too much. Of course, if you forget things all the time, like really important things, then maybe it’s a good idea to talk to someone about it. But for the most part, forgetting things is just a normal part of being human. Sometimes, I forget things and then remember them later, like in the middle of the night or when I’m doing something completely unrelated. It’s funny how that works. The other day, I forgot where I put my keys, and I looked everywhere for them, and then I found them in my pocket. That was pretty funny, and I laughed about it. Forgetting things can be annoying, but it can also be kind of funny, especially when you realize how silly it is. Sometimes, I forget things because I’m distracted or because I’m thinking about something else, or because I’m just not paying attention. I think paying attention is really important if you want to remember things, but it’s hard to pay 

\end{outputbox}

\newpage
\subsection{Prefix Conditioned Samples}
\textbf{Example Generation 1}
\begin{outputbox}
   
    {
    \textcolor{blue}{
    The history of artificial intelligence
    }
    }
      is a fascinating journey that spans multiple decades, encompassing a series of breakthroughs, setbacks, and paradigm shifts. In the mid-twentieth century, the groundwork for AI was laid by visionary thinkers who dreamed of machines that could mimic human intelligence. Alan Turing, often regarded as the father of theoretical computer science, proposed the idea of a machine that could think and introduced the Turing test as a measure of intelligence. Early AI research was characterized by optimism and bold predictions. Scientists and mathematicians believed that within a few years, computers would be capable of complex reasoning, natural language understanding, and even creative thought. The initial wave of AI, known as symbolic AI, relied on rule-based systems and logic. Researchers painstakingly encoded human knowledge into computer programs, hoping to create systems that could solve problems, play games, and answer questions. The field experienced its first wave of enthusiasm in the 1950s and 1960s, with projects such as the General Problem Solver and the development of the first expert systems. However, as researchers soon discovered, human intelligence is far more complex than initially imagined. The limitations of symbolic AI became apparent as systems struggled to handle ambiguity, context, and real-world variability. The field entered a period known as the "AI winter," where funding and interest declined due to unmet expectations. Despite these setbacks, research continued behind the scenes, and new approaches began to emerge. The advent of machine learning in the 1980s and 1990s marked a significant turning point. Instead of relying on hand-coded rules, researchers began to develop algorithms that could learn from data. Early machine learning techniques, such as decision trees and neural networks, showed promise but were limited by the available computational power and data. The rise of the internet and the digital revolution in the late 1990s and early 2000s provided the necessary infrastructure and data for machine learning to flourish. With the explosion of online information, researchers had access to vast amounts of data that could be used to train more sophisticated models. The development of deep learning in the 2010s further accelerated progress in AI. Deep neural networks, inspired by the structure of the human brain, demonstrated remarkable capabilities in image and speech recognition, natural language processing, and even game playing. Breakthroughs such as AlphaGo, which defeated world champions in the ancient game of Go, captured the public imagination and demonstrated the potential of AI to surpass human performance in specific domains. Today, artificial intelligence is deeply integrated into many aspects of daily life, from virtual assistants and recommendation systems to autonomous vehicles and medical diagnostics. AI technologies are transforming industries, enabling new forms of creativity, and raising important ethical and societal questions. As AI continues to evolve, researchers are exploring new frontiers, such as explainable AI, which aims to make machine learning models more transparent and interpretable. The future of artificial intelligence holds both promise and challenges. On one hand, AI has the potential to solve some of the world's most pressing problems, from climate change to healthcare. On the other hand, concerns about job displacement, privacy, and the misuse of AI remain significant. The ongoing development of artificial intelligence will require collaboration across disciplines, thoughtful regulation, and a commitment to ethical principles. As we look ahead, it is clear that AI will play an increasingly central role in shaping the future of humanity, driving innovation and transforming the way we live, work, and interact with technology. The next decades will undoubtedly bring new breakthroughs, challenges, and opportunities, as artificial intelligence continues to push the boundaries of what is possible. The journey of AI is far from over, and the story of its evolution is still being written. Every day, researchers around the world are working to unlock new capabilities, address limitations, and ensure that artificial intelligence is developed and deployed in ways that benefit society as a whole. From its humble beginnings to its current state of rapid advancement, artificial intelligence represents one of the most exciting and transformative fields in science and technology. Whether AI will ever achieve true general intelligence or consciousness remains an open question, but the progress made so far is a testament to human ingenuity and the power of collaboration. As we continue to explore the possibilities of artificial intelligence, we must remain mindful of the ethical implications and strive to create a future where AI serves the greater good. The evolution of artificial intelligence is not just a story of technology, but also a story of human ambition, creativity, and the endless pursuit of knowledge.
\end{outputbox}

\newpage

\textbf{Example Generation 2}

\begin{outputbox}
    
    {
    \textcolor{blue}{
    Today the cold war ended.
    }
    }
     Or at least, that’s what they say. I’m not sure if it really ended, or if it just changed, or maybe it’s still going on in some way. I think the Cold War was about the United States and the Soviet Union, and they were always arguing or threatening each other, but they never actually fought a big war. Or maybe they did, but not the kind of war with soldiers and tanks, more like a war of words and spies and maybe space. I remember something about the space race, where they tried to see who could get to the moon first. Or was it Mars? Maybe it was the moon. I think the United States got to the moon first, but I’m not sure. Anyway, today the Cold War ended, or at least that’s what the news said. I don’t know if it’s true, because sometimes the news gets things wrong, or maybe they just want people to feel better. I think the Cold War was about nuclear weapons, too, and everyone was scared that someone would push the button and start a big explosion. I remember hearing about bomb shelters and people practicing hiding under their desks, which seems silly now, but maybe it made sense back then. I’m not sure. Today, the Cold War ended, but I don’t know what that means for the world. Maybe things will be better now, or maybe they’ll just be different. I think the Cold War lasted a long time, like maybe forty years, or maybe more. I’m not good with dates, so I might be wrong. I think it started after World War II, but I’m not sure. I remember something about the Berlin Wall, and how it fell, and people were happy, but I’m not sure if that was the end of the Cold War or just a part of it. Today, the Cold War ended, but I don’t know if anyone will notice. I think most people are busy with other things, like work or school or watching TV. Maybe the Cold War ending is a big deal, but maybe it’s not. I’m not sure. I think the Cold War was about politics and power, and maybe about who could control the world, or at least parts of it. I remember something about communism and capitalism, and how they were fighting over which one was better. I’m not sure which one is better, but I think most people like capitalism, at least where I live. I think the Cold War was also about movies and music, and how each side tried to show that their way of life was better. I remember something about jazz and rock and roll, and how the Soviet Union didn’t like it, but I’m not sure if that’s true. Today, the Cold War ended, but I don’t know what will happen next. Maybe there will be more peace, or maybe there will be new problems. I think the world is always changing, and it’s hard to keep up. I think the Cold War was a big part of history, and maybe it will be in the history books, but I’m not sure if kids will care. I think most people just want to live their lives and not worry about big things like wars or politics. Today, the Cold War ended, but I’m not sure if it really matters. I think it’s good to have peace, but I don’t know if that’s what will happen. Maybe there will be new wars, or maybe things will stay the same. I’m not sure. I think the Cold War was about fear, and about people being scared of each other, and maybe that’s still happening, even if the Cold War is over. I think fear is a big part of life, and maybe it will always be there, no matter what happens. Today, the Cold War ended, but I don’t know if that will change anything. I think the world is complicated, and it’s hard to understand everything. I think the Cold War was about more than just two countries, but I’m not sure what else it was about. Maybe it was about ideas, or about the future, or about who gets to decide what happens next. I’m not sure. Today, the Cold War ended, but I don’t know if that’s true, or if it’s just something people are saying. I think the world is always changing, and it’s hard to keep up. I think the Cold War was a big part of history, and maybe it will be in the history books, but I’m 

\end{outputbox}

\newpage

\textbf{Example Generation 3}

\begin{outputbox}
    
    {
    \textcolor{blue}{
    Jazz music is really something special.
    }
    }
     Or at least that’s what I think. I’m not sure where jazz started, but I think it was in America, maybe in New Orleans, or maybe somewhere else. I remember hearing something about jazz being a mix of different kinds of music, like blues and ragtime and maybe even classical, but I’m not sure. Jazz is kind of hard to explain, but it’s all about improvising, or making things up as you go along. Sometimes jazz musicians just start playing and see where the music takes them, and that’s really cool. I think jazz is about freedom, and about expressing yourself, and about not following all the rules. I’m not sure if that’s true, but it seems like jazz is always changing, always moving, always trying new things. I remember hearing jazz on the radio when I was a kid, and it always made me feel happy, or maybe a little sad, depending on the song. I think jazz can make you feel all kinds of things, and that’s part of what makes it special. Jazz is also about the instruments, like the saxophone and the trumpet and the piano, and sometimes the drums and the bass. I remember seeing a jazz band once, and the saxophone player was amazing, just playing all these crazy notes and making the music come alive. Sometimes jazz musicians play solos, where one person gets to shine, and everyone else supports them, and that’s really nice. I think jazz is about community, and about sharing something special with other people. I’m not sure if that’s true, but it seems like jazz brings people together, even if they don’t know each other. I just know that jazz is always evolving, always trying new things, and that’s part of what makes it exciting. I think jazz is about taking risks, and about not being afraid to make mistakes, because sometimes the best music comes from mistakes. I remember hearing a jazz musician say that, and it made sense to me. I think that’s why jazz is so powerful, because it comes from the inside, from the soul, or at least that’s what people say. I’m not sure if that’s true, but it sounds nice. Jazz is also about rhythm, and about the beat, and about making people want to move. I think jazz can make you dance, or at least tap your foot, and that’s always fun. I remember hearing jazz at a party once, and everyone was dancing and having a good time, and it was really cool. Jazz is also about the night, and about staying up late and listening to music, and about feeling alive. I think jazz is best at night, when everything is quiet and you can really listen. I’m not sure if that’s true, but it seems like jazz is made for the night. Jazz is also about the past, and about remembering all the great musicians who played before, like Louis Armstrong and Miles Davis and Duke Ellington, but I’m not sure if I got all the names right. I think jazz is about tradition, but also about breaking the rules and trying new things. I’m not sure if that makes sense, but it seems like jazz is always changing, always moving, always trying to find new ways to express itself. I think jazz will always be around, because it’s so flexible and so open to new ideas. I’m not sure if that’s true, but I hope it is. Jazz is also about listening, and about paying attention, and about really hearing the music. I think jazz is best when you listen closely, and when you let yourself get lost in the sound. I remember hearing jazz on the radio, and sometimes I would just close my eyes and listen, and it felt like the music was taking me somewhere else. Jazz is also about mistakes, and about not being perfect, and about being okay with that. I think jazz is about taking chances, and about not being afraid to try something new, even if it doesn’t work out. I remember hearing a jazz musician say that, and it made sense to me. Jazz is also about feeling, and about putting your heart into the music, and about letting go and just playing. I think that’s why jazz is so powerful, because it comes from the soul. I’m not sure if that’s true, but it sounds nice. Jazz is also about rhythm, and about the beat, and about making people want to move. I think jazz can make you dance, and that’s always fun. I remember hearing

\end{outputbox}

\section{GPT-style Energy Functions}
Given a GPT-style causal generative model with parameters $\theta$, we consider the generative perplexity of this model, i.e., $PPL_\theta (x) = \frac{1}{L}\sum\limits_{i=1}^{L-1} \log p_\theta (x_{i+1} \vert x_{1:i})$ as our energy function. In any iteration then, our forward process essentially has to take some index, say $i$ and update $x_i$ to some other token $y$ with probability  $p(x_i = y_i \ \vert x_{\setminus i}) = \min\left\{1,e^{-PPL_\theta(x_{\setminus i}, y_i) + PPL_\theta(x)}\right\}$ for $y_i \in \Sigma$ (possibly with some self-loops for lazy chains). This ensures that the stationary distribution of the Markov chain is the distribution of the GPT-style causal generative model itself. Given this setup, we can then reverse our Markov process by learning (time-varying) jump probabilities using this transition kernel and stationary distribution and feeding it into a denoising score entropy loss function (\ref{eqn:dwdsel}). However, we will have to keep a set of parameters for this GPT-style model for our energy function and need additional parameters for the score network required in the reverse process. From a statistical viewpoint, it would then only be fair to compare such models with say GPT-style models with parameter count equaling the parameter count of the GPT-model used for energy PLUS the one used for the denoising score network. Furthermore, we would have to train the denoising score network from scratch which can be costly. 

As we will show in our second approach based on UL2 style models, we can devise a strategy where essentially the number of model parameters are roughly unchanged and the training can also be set up to mostly look like fine-tuning a model on a score entropy loss function. It is also important at this point to remark that to execute each forward transition in the GPT-style Energy Function Glauber Dynamics, we would need to compute $|\Sigma|$ many perplexity evaluations, one for each $y_i$ which can be quite costly but note that all these perplexity calculations can be evaluated in parallel. However, the computational time overhead can still be significant. It is still worth considering this approach to compare against other methods provided one has enough computational resources and we leave this comparison for future work. 

\section{Training Hyperparameters}
We use the T5 tokenizer which has a vocabulary of 32100 and we train on sequences of length 1024. We use the AdamW optimizer (with $\beta_1 = 0.9, \beta_2 = 0.999, \varepsilon = 10^{-8}$ and 0.9999 EMA, no weight decay and a linear warmup schedule for the first 9000 steps to $3 \times 10^{-4}$ and then decay to $10^{-6}$ using a cosine decay schedule over the remaining steps and we use a gradient accumulation of 4. We train our models using 32 H100s for 2 epochs which takes approx 6 days to finish training for our larger model (we highlight that based on the reported training time of the GGM paper \citep{varma2024glauber}, their training takes about 8 days to train a model one size smaller than ours using TPU compute which on a TFLOP basis is roughly equivalent to 24 H100s). We use 128 examples in each batch with 32 different time-steps bringing our batch size to 4096 effectively.

\newpage

\end{document}